\documentclass[journal]{IEEEtran}
\usepackage{graphicx}
\usepackage{enumerate}
\usepackage{amsmath,amsfonts,amssymb}
\usepackage{color}
\usepackage{multirow}
\usepackage[noend]{algpseudocode}
\usepackage{algorithm}
\usepackage{algpseudocode}
\usepackage{setspace}
\usepackage{lineno,hyperref}
\usepackage{booktabs}
\usepackage[table]{xcolor}
\modulolinenumbers[5]
\usepackage{etoolbox}
\makeatletter
\patchcmd{\@makecaption}
  {\scshape}
  {}
  {}
  {}
\makeatletter
\patchcmd{\@makecaption}
  {\\}
  {.\ }
  {}
  {}
\makeatother
\begin{document}
%\begin{frontmatter}

%\title{Distribution-aware Interactive-Attention Deep Neural Network for Cloud Recognition in Transformed FY-4A Satellite Image}
%\renewcommand{\thefootnote}{\fnsymbol{footnote}}
%\author{Weiying Xie$^{1}$\footnote{Corresponding author: zwpxwy@126.com or wyxie@xidian.edu.cn (Weiying Xie)}, Jiaqing Zhang$^{1}$, Jie Lei$^{1}$, Yunsong Li$^{1}$, Xiuping Jia$^{2}$,}
%\address{$^{1}$State Key Laboratory of Integrated Services Networks, Xidian University, Xi'an 710071, China}
%\address{$^{2}$School of Engineering and Information Technology, The University of New South Wales, Canberra, ACT 2600, Australia}
\title{Guided Hybrid Quantization for Object Detection in Multimodal Remote Sensing Imagery via One-to-one Self-teaching}

\author{Jiaqing Zhang,
	Jie Lei,~\IEEEmembership{Member,~IEEE},
	Weiying Xie,~\IEEEmembership{Member,~IEEE},
%	Zhenman Fang,~\IEEEmembership{Member,~IEEE},
	Yunsong Li,~\IEEEmembership{Member,~IEEE},
	and Xiuping Jia, ~\IEEEmembership{Fellow, IEEE}
%	and Qian Du,~\IEEEmembership{Fellow,~IEEE}
%	\thanks{This work was supported in part by the National Natural Science Foundation of China under Grant 62071360, Grant 61571345, Grant 91538101, Grant 61501346, Grant 61502367 and Grant 61701360, in part by the Young Talent fund of University Association for Science and Technology in Shaanxi of China under Grant 20190103, in part by the Special Financial Grant from the China Postdoctoral Science Foundation under Grant 2019T120878, in part by the 111 project under Grant B08038, in part by the Fundamental Research Funds for the Central Universities under Grant XJS200103, in part by the Natural Science Basic Research Plan in Shaanxi Province of China under 2019JQ153, Grant 2016JQ6023, and Grant 2016JQ6018, in part by the General Financial Grant from the China Postdoctoral Science Foundation under Grant 2017M620440, in part by the Yangtse Rive Scholar Bonus Schemes under Grant CJT160102, in part by the Ten Thousand Talent $(Corresponding~authors:)$
\thanks{This work was supported in part by the National Natural Science Foundation of China under Grant 62071360.

J. Zhang, J. Lei, W. Xie, Y. Li are with the State Key
Laboratory of Integrated Services Networks, Xidian University, Xi’an 710071,
China (e-mail: jqzhang\underline{ }2@stu.xidian.edu.cn; jielei@mail.xidian.edu.cn; wyxie@xidian.edu.cn; ysli@mail.xidian.edu.cn).

%Z. Fang is with School of Engineering Science, Simon Fraser University, Burnaby, BC, Canada (e-mail: zhenman@sfu.ca).

%Q. Du is with the Department of Electronic and Computer Engineering, Mississippi State University, Starkville, MS 39759 USA (e-mail: du@ece.msstate.edu).

X. Jia is with the School of Engineering and Information Technology, The University of New South Wales, Canberra, ACT 2600, Australia (e-mail: xp.jia@ieee.org).

}

}
%IEEE Transactions on Neural Networks and Learning Systems
\markboth{IEEE TRANSACTIONS ON GEOSCIENCE AND REMOTE SENSING,~Vol.~X, No.~X, X~2022}%
{Zhang \MakeLowercase{\textit{et al.}}: Guided Hybrid Quantization for Object detection in Multimodal Remote Sensing Imagery via One-to-one Self-teaching}

\maketitle

% !TeX spellcheck = en_US
\begin{abstract}
Recently, deep convolution neural networks (CNNs) have promoted accuracy in the computer vision field. However, the high computation and memory cost prevents its development in edge devices with limited resources, such as intelligent satellites and unmanned aerial vehicles. Considering the computation complexity, we propose a \underline{G}uided \underline{H}ybrid Quantization with \underline{O}ne-to-one \underline{S}elf-\underline{T}eaching (\textbf{GHOST}) framework. More concretely, we first design a structure called guided quantization self-distillation (GQSD), which is an innovative idea for realizing lightweight through the synergy of quantization and distillation. The training process of the quantization model is guided by its full-precision model, which is time-saving and cost-saving without preparing a huge pre-trained model in advance. Second, we put forward a hybrid quantization (HQ) module to obtain the optimal bit width automatically under a constrained condition where a threshold for distribution distance between the center and samples is applied in the weight value search space. Third, in order to improve information transformation, we propose a one-to-one self-teaching (OST) module to give the student network a ability of self-judgment. A switch control machine (SCM) builds a bridge between the student network and teacher network in the same location to help the teacher to reduce wrong guidance and impart vital knowledge to the student. This distillation method allows a model to learn from itself and gain substantial improvement without any additional supervision. Extensive experiments on a multimodal dataset (VEDAI) and single-modality datasets (DOTA, NWPU, and DIOR) show that object detection based on GHOST outperforms the existing detectors. The tiny parameters ($<$9.7 MB) and Bit-Operations (BOPs) ($<$2158 G) compared with any remote sensing-based, lightweight or distillation-based algorithms demonstrate the superiority in the lightweight design domain. Our code and model will be released at \url{https://github.com/icey-zhang/GHOST}.
\end{abstract}
% !TeX spellcheck = en_US
\section{Introduction}
%\begin{figure}[htb]
%	\centering
%	\includegraphics[scale=0.48]{./Figures/quantization.pdf}
%	\centering
%	\caption{.}
	%\vspace{-0.1in}
%	\label{quantization}
%\end{figure}

%\begin{figure*}[htb]
%	\centering
%	\includegraphics[scale=0.85]{./Figures/frequencyfusion.png}
%	\centering
%	\vspace{-0.05in}
%	\caption{(a)(b)}
%	\vspace{-0.1in}
%	\label{frequencyfusion}
%\end{figure*}

\IEEEPARstart{O}{bject} detection in aerial images plays an important role in military security aiming to locate interested objects (e.g., vehicles, airplanes) on the ground and identifying their categories \cite{8953881}. From universal detectors for natural images such as YOLOv3 \cite{redmon2018yolov3}, Faster R-CNN \cite{ren2015faster}, FCOS \cite{tian2019fcos} are widely introduced in the field of remote sensing (RS); more and more dedicated detectors for RS scene are designed and improved with the requirements of objects tasks.
%Convolution neural networks (CNNs) are widely adopted for detection with the development of deep learning and achieving state-of-the-art performances on various benchmarks. 
%recent years have witnessed the vast appearance of a large amount of remote sensing satellite data because the number of satellite launches has exploded. 
%However, massive multimodal images are collected from satellites, drones, and airplanes with the sensor technology flourishing which brought heavy pressure on data transmission and processing. 
However, the large complexity of the object detection network is under-investigated, which limits the practical deployment under resource-limited scenarios and bring a heavy burden to process massive multimodal images collected from satellites, drone, and airplanes. %Recently, more light is shed on the design of lightweight detectors to address this problem. 
Hence, a series of compression schemes have been proposed to settle this problem, such as pruning \cite{9157802}, quantization \cite{li2019fully,wang2021generalizable,zhou2016dorefa} and distillation \cite{wang2021distilling, dai2021general}. 

\begin{figure}[htb]
	\centering
	\includegraphics[scale=0.43]{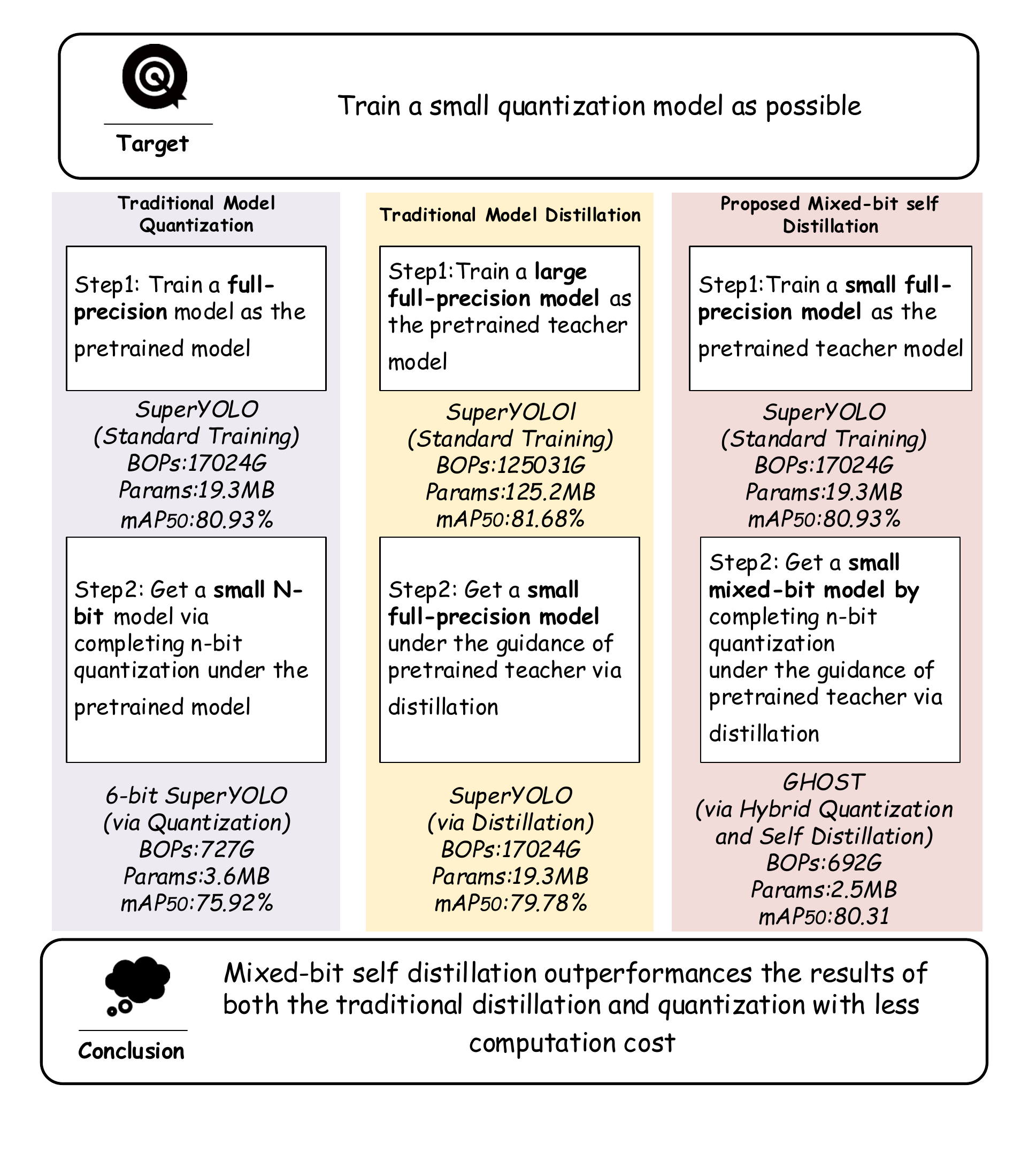}
	\centering
	\caption{Comparison of training complexity, and accuracy between traditional distillation, traditional quantization and proposed mixed-bit self distillation (reported on VEDAI).}
	%\vspace{-0.1in}
	\label{compare}
\end{figure}

%to relieve the computational cost of deep models on the lightweight detectors, e.g., LO-DET  \cite{9390310}. %Some issues such as weak feature representation ability, high false alarm, and imprecise regression bounding boxes \cite{yang2022adaptive} still remain. 
Quantization algorithms \cite{yang2021bsq, liu2020reactnet} directly compress the cumbersome network, effectively reducing the computation cost and model size with a great compression potential. However, trivially applying quantization to CNNs usually leads to inferior performance if the compression bit decreases to a low level. Some knowledge distillation methods \cite{hinton2015distilling, gou2021knowledge, yang2022adaptive} are proven to be valid to elevate the performance of the lightweight model but have to pre-train a huge teacher model as a guidance of the student model which is time-consuming and resource-consuming \cite{2019Be}. Self-distillation methods \cite{ji2021refine, 2019Be, hou2019learning} overcome this problem via the transfer of information inside the model itself without introducing extra huge storage and time consuming from the teacher model.
%量化是一个具有很大压缩潜力的一种压缩方式，但是会导致较差的性能当压缩比特变为一个很小的数量级时。

The above view naturally leads to a question: \textbf{What research results will we get if we combine quantization and distillation by using a small full-precise network to guide the learning process of a quantization for this full-precision network?} In this way, the tremendous compression capacity of quantified networks and the performance of full precision networks can be collaborative and cooperative.

In this paper, we design an adaptive one-to-one educational policy pertaining the full-precision network and the quantization network. We propose a simple yet novel approach that allows quantization network to reinforce presentation learning of itself relative full-precision network without the need of additional labels and external supervision. Our approach is named as \textit{Guided Hybrid Quantization with One-to-one Self-Teaching} (GHOST) based on the guided quantization self-distillation (GQSD) framework. As the name implies, GHOST allows a network to exploit useful and vital knowledge derived from its own full-precision layers as the distillation targets for its quantization layers. GHOST opens a new possibility of training accurate tiny object detection networks.

As shown in Fig. \ref{compare}, in order to train a small compact model to achieve as high accuracy as possible with less computation cost, we propose mixed-bit self distillation framework. Instead of implementing two steps in traditional distillation, which means that to train a large teacher model comes first, following by distilling the knowledge from it to the student model, we propose a two-step mixed-bit self distillation framework, in which the training process of the second quantization step is based on the pretrained small full-precision model. The proposed framework not only requires less computation cost (from 20797 G BOPs to 692 G BOPs on VEDAI dataset, a 30X faster training cost), but also can accomplish much higher accuracy (from 75.84\% in traditional quantization to 80.31\% on SuperYOLO)
The main contributions of our work are as follows:

\begin{itemize}
    \item We propose a unified guided quantization thought based on self-distillation called GQSD, which can tackle the lightweight object detectors' quantization optimization problem in remote sensing. We are the first to formulate an adaptive one-to-one education policy between the full-precision network and the quantization network at the same structure in object detection.
    \item For the finding of weight value distribution features of remote sensing images, we design a hybrid quantization module, whose adaptive selection of the core information of the weights for quantization with a constrained preset condition can keep the balance of accuracy and efficiency.
    \item Aiming to offset the loss of the quantization information, the switch control machine is adopted to enable the student to distinguish and close the teacher's wrong guidance and mine the correct and vital knowledge from self-distillation. 
    
\end{itemize}

The rest of this paper is organized as follows: Section \ref{sec: Related Work} give a rough overview of the spacific related work to this paper. Section \ref{sec: Network Architecture} presents our proposed method in detail. Section \ref{sec: Experiment} introduces experimental results and analysis. Section \ref{sec: Conclusion} concludes this paper and discusses the future work.

%\begin{figure*}[htb]
%	\centering
%	\includegraphics[scale=0.9]{./Figures/framework.png}
%	\centering
%	\caption{The overview of the proposed SuperYOLO framework.}
	%\vspace{-0.1in}
%	\label{framework}
%\end{figure*}
% !TeX spellcheck = en_US
\section{Related Work}
\label{sec: Related Work}
In this section, we reviewed related work from object detection and network compression and acceleration in detail.  

\begin{figure*}[htpb]
	\centering
	\includegraphics[scale=0.9]{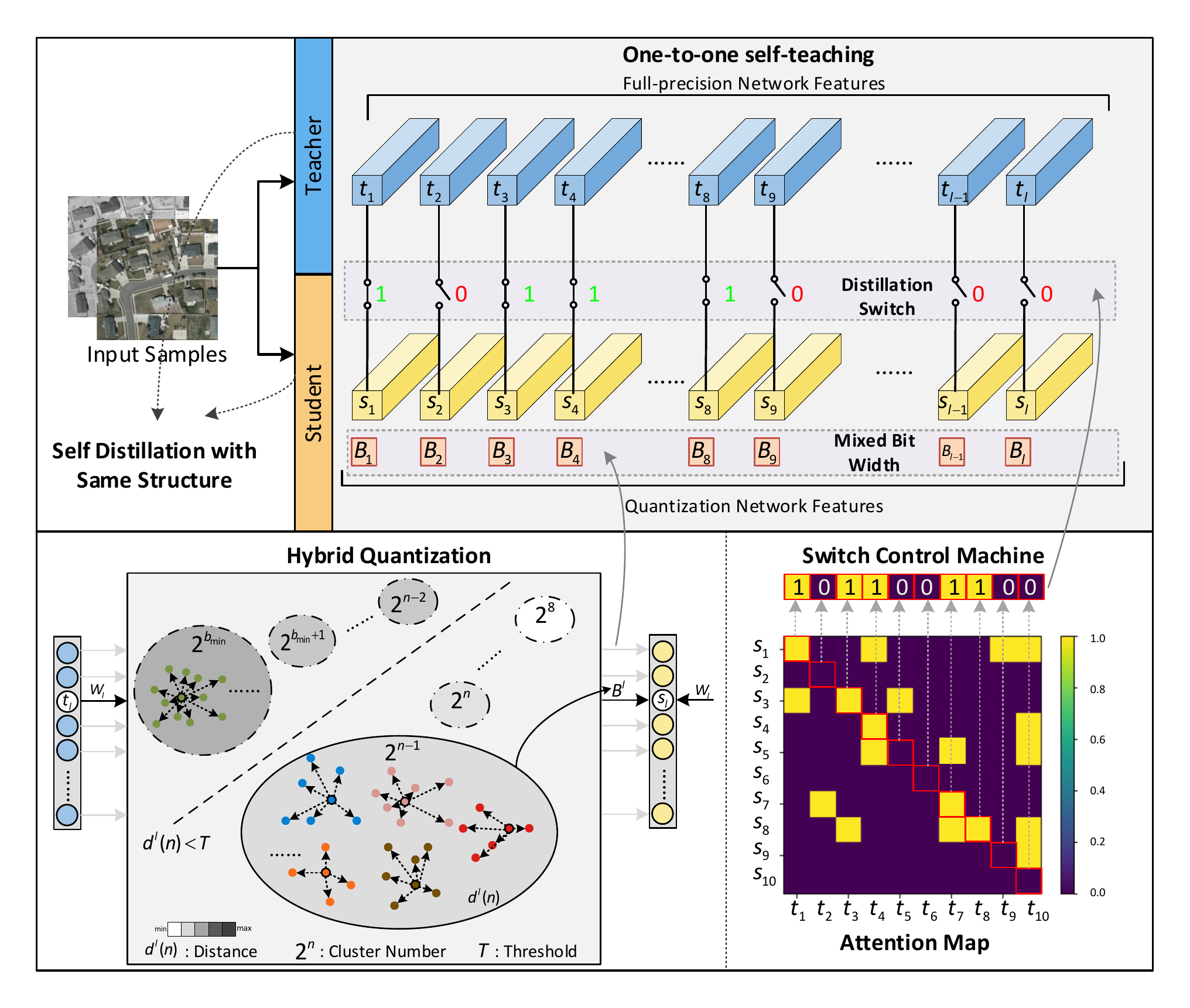}
	\centering
	\caption{Overview of our proposed framework. An attention-based model determines similarities between the teacher and student features. Knowledge from each teacher feature is transferred to the student with similarities identified by Switch control machine (SCM) by self-distillation with the same structure. The mixed bit widths of the student network for quantization are based on the search results of the full-precision weights research space of teacher network in the same layer.}
	%\vspace{-0.1in}
	\label{frame}
\end{figure*}

\subsection{Object Detection with Deep Learning}
Various CNN-based object detection architectures have shown promising performance, bringing the field to a new level. The architectures can be roughly divided into two main domains: two-stage, and one-stage according to the change process of proposals.

\textbf{Two-stage Detectors:} A typical method is selective search work \cite{uijlings2013selective}, where the first stage is to generate a large set of proposed region candidates that are required to cover the whole objects and then filter out most negative positions, and then the second stage is to complete classification for each region. 
R-CNN \cite{6909475} creates a new era as one of the most successive two-stage algorithms owing to the upgrading of the second-stage classifier to a convolution network.  Fast-RCNN \cite{7410526} extractes features over the images before proposing regions and integrates the extractor and classifier by employing a soft layer rather than SVM classifier. Faster R-CNN \cite{ren2015faster} introduces a CNN-based region proposal network to further integrate proposal generation with the second-stage classifier into a single convolution network.

\textbf{One-stage Detectors:} One-stage detectors aim to jointly predict the classification and location of objects by integrating the detection and classification process. Recently, a series of SSD \cite{liu2016ssd, fu2017dssd, cui2018mdssd} and YOLO \cite{redmon2016you, redmon2017yolo9000, redmon2018yolov3, bochkovskiy2020yolov4} have renewed interest in the one-stage object detection. SSD implements independent detection on multiscale feature maps, while the YOLO utilizes combined detection. These methods have paid more attention to speed, but their accuracy trails behind that of tow-stage methods. YOLOv2 \cite{redmon2017yolo9000} modifies the location regression pattern depending on bounding boxes. YOLOv3 \cite{redmon2018yolov3} considers multiscale objects and detects them in the three scales, which can realize the detection of multiple sizes of objects. YOLOv4 \cite{bochkovskiy2020yolov4} introduces more data augmentation tricks, activation functions, backbone structures, and IoU loss metrics to enhance the robustness of the network. YOLOv5 \cite{yolov5} releases four different size models, where the basic structures are identical, which allows YOLOv5 to have higher flexibility and versatility in practical applications. To solve the dilemma of the category imbalance, RetinaNet \cite{lin2017focal} reduces the weight of massive amounts of simple negative samples in training by designing a focal term for cross-entropy loss. As FCOS \cite{tian2019fcos} which belongs to anchor-free methods is proposed, adjusting hyperparameters and calculations related to anchor boxes has been avoided. ATSS \cite{zhang2020bridging} selects positive samples adaptively to enhance the detection performance.
\subsection{Deep Network Compression and Acceleration}
% Knowledge Distillation (KL) is a method of transferring knowledge from a pretrained teacher network to a student network, so a smaller network can replace a large teacher network at the deployment stage. Previous methods mostly focus on proposing feature transformation. 
Although the speed of the one-stage detection network is superior, its large model and high computation complexity still deserves to explore. Some researches focus on the design of a lightweight backbone. MobileNetV2 \cite{sandler2018mobilenetv2} utilizes the depthwise separable convolutions to build a lightweight model. ShuffleNet \cite{zhang2018shufflenet}, and SqueezeNet \cite{iandola2016squeezenet} also effectively reduces the memory footprint during inference and speed up the detection. In the literature, a potential direction of model compression is knowledge distillation (KD) which concentrates on transferring knowledge from a heavy model (teacher) to a light one (teacher) to improve the light model's performance without introducing extra costs \cite{zhao2022decoupled, zhang2021learning}. Whereas the knowledge distillation enables utilizing the larger network in a condensed manner, the pretraining of the large network requires extra substantial computation resources to prepare the teacher network \cite{ji2021refine}. The preparation of the pretrained teacher network is time-consuming and cost-consuming. The self-knowledge distillation \cite{ji2021refine, 2019Be, hou2019learning} can overcome this problem by distilling its own knowledge without prior preparation of the teacher network. Quantization is another way to compact the model directly and compress the ponderous network by using low-bit representation.  Mixed-precision quantization method uses different numbers of bits for a given data type to represent values in weight tensor. Many works \cite{yang2021bsq, cai2020zeroq, dong2020hawq, dong2020hawq} have shown that the mixed-precision method is efficient for quantizing network layers that have different importance and sensitiveness for the bit width.  However, trivially applying quantization to CNNs usually leads to inferior performance if the compression bit decreases to a low level.

% !TeX spellcheck = en_US
\section{Network Architecture}
\label{sec: Network Architecture}
In this section, we first revisit conventional KD and describe the proposed GHOST framework in Sec. \ref{sec: Overview}. Then, we present the details of the inspired hybrid quantization algorithm (Sec. \ref{sec: Quantization}) and this quantization training process is guided by a one-to-one self-teaching method illustrated in Sec. \ref{sec: Distillation}.

\subsection{Overview}
\label{sec: Overview}
%Relational Knowledge Distillation
%Knowledge distillation is widely applied in classification for nature images xxx. 
KD is a widely-applied method that can be expressed as a knowledge transformer from teacher to student. Given a teacher model $\rm T$ and a student model $\rm S$, the $x$ is the data examples of models, here they can be the same for the teacher and the student model. In general, the KD machine can be uniformly expressed as:
\begin{equation}
	\min {{\mathcal{L}}_{KD}} = \min \sum\limits_{{x_i} \in x} {\mathcal{L}} \left( {{\rm{T}}\left( {{x_i}} \right),{\rm{S}}\left( {{x_i}} \right)} \right),
\end{equation}
where $\mathcal L$ is the loss function that penalizes the differences between the teacher and the student. 

The student model size is commonly designed in a small size to achieve the purpose of model compression in which the performance of the student can chase the teacher but consumes a computing-friendly resource. Nonetheless, the computation cost of the student model is larger than the pruning method directly completed on the teacher model. This demonstrates that the existing KD-based quantization algorithms still have great potential room for improvement.

We aim at developing a novel and generic baseline network with a focus on the learnable knowledge characteristics, making it well-applicable to the highly accurate and fine object detection of RS images with less computational costs. The key to model quantization with knowledge learning is to reduce the discrepancy which can be punished by distance or angle loss function between full-precision model $\rm P$ (teacher) and low-precision model $\rm Q$ (student) through optimizing $\rm Q$, which can be expressed as: 
\begin{equation}
	\mathrm{Q^*} = \min_{\mathrm{Q}} \sum_{x_i \in x} \mathcal{L}({\mathrm{P}}(x_i),{\mathrm {Q}}(x_i)).
\end{equation}
The weights of the teacher are frozen without gradient propagation when the teacher network guides the training of the student network.
Based on the above presentation, we design an effective teacher-student distillation framework called GQSD which can be represented as:
%Category Correlation and Adaptive Knowledge Distillation for Compact Cloud Detection in Remote Sensing Images
%We define the behavior of the teacher or student as two parts: the constructive knowledge from the feature map $F$ and the label output $O$ which can be represented as:
%\begin{equation}
%t\in({G(F^t),O^t}), \quad s\in({G(F^s),O^s}),
%\end{equation}
%where $G$ is a relational potential function conceived to extra internal relationship from feature map F. The final optimization can be defined as:
%\begin{equation}
%\min \mathcal L_{KD} = \mathop {\min } (\sum_{x_i \in x} \mathcal L_F(G(F^t),{\rm{ }}G(F^s))+ \mathcal L_O(O^t,{\rm{ }}O^s)).
%\end{equation}
\begin{equation}
	\begin{split}
	\min \mathcal L_{KD} &= \min \sum\limits_{{x_i} \in x} \mathcal L ({\rm P}({x_i}),{\rm Q}({x_i}),{\rm R}({x_i})), \\
	s.t. &\quad W_Q=W_{init}, B_Q=B.
	\end{split}
\end{equation}
%全精度网络作为教师指导模型的量化过程
Specifically, the full precision network is the import fundamental teacher which not only provides the initial weights $W_{init}$ and bit width $B$ of the quantization model but also guides the quantization process to mine the vital knowledge from the teacher in specific features selected by a control ${\mathrm{R}}$. As shown in Fig \ref{frame}, we propose a GHOST framework that concludes a hybrid quantization (HQ) module and a one-to-one self-teaching (OST) module. The mixed bit widths of the student network for quantization are based on the search results of the full-precision weights research space of the teacher network in the same layer. Inspired by the idea of harnessing intermediate features to improve performance in knowledge distillation \cite{liu2021zero,ji2021show,chen2021distilling}, we design a Switch Control Machine (SCM) as ${\mathrm{R}}$ to generate an attention map that gains intermediate feature similarities between the teacher and student. The SCM controls the distillation switch and determines which knowledge should be delivered dynamically. Knowledge from each teacher feature is transferred to the student with similarities identified by SCM by self-distillation with the same structure. With a pretrained full-precision model as a initial weight, the quantization and distillation processes are conducted simultaneously to ultimately obtain a small lightweight model with little loss of accuracy. The details of the modules will be described separately as follows.

\begin{figure*}[htb]
	\centering
	\includegraphics[scale=0.6]{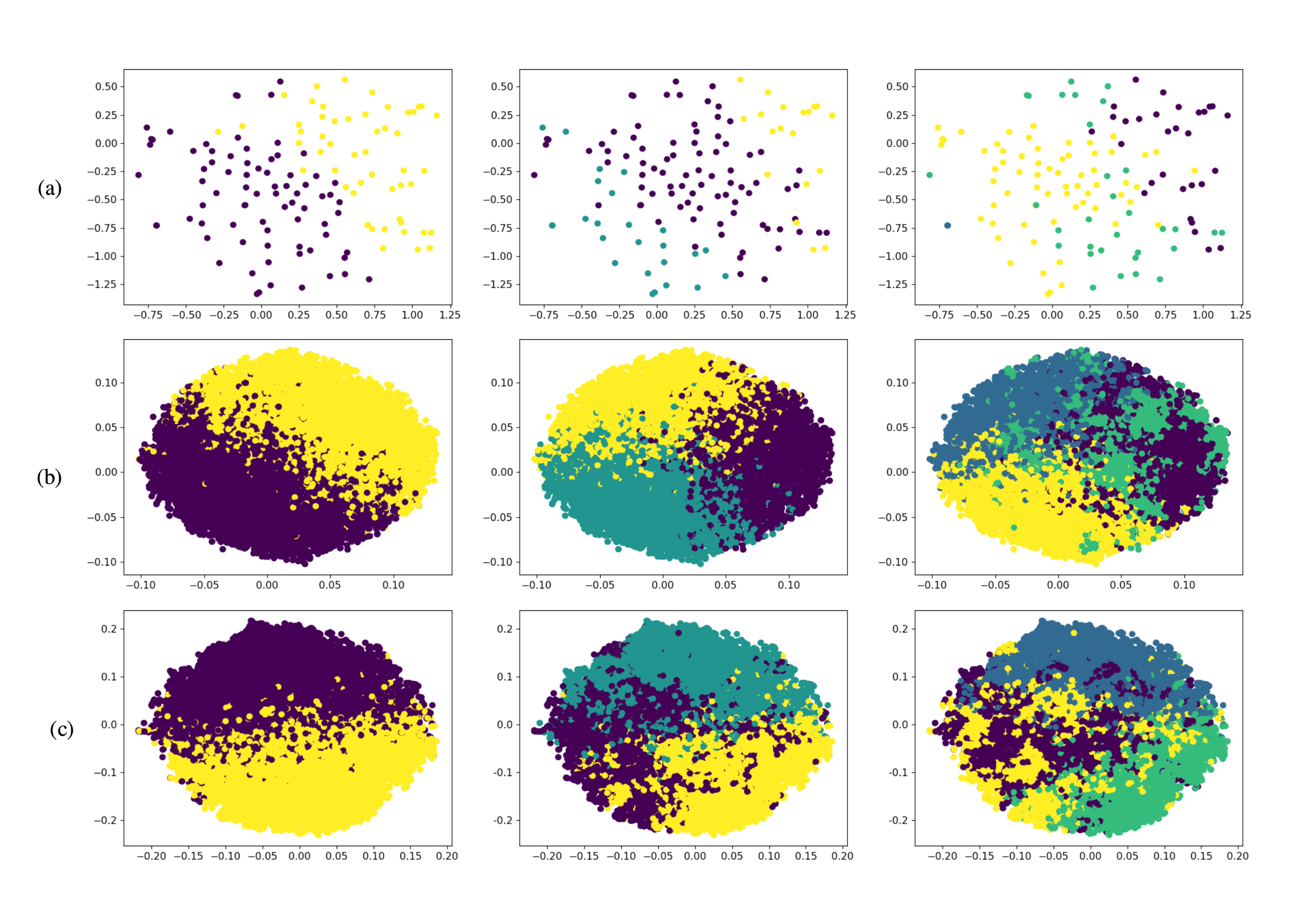}
	\centering
	\caption{In the bottom layer of a trained network, the feature distributions of different categories intervene with each other severely as (a) shows. Many delicate neurons are needed to distinguish the overlapped distributions. And as the network propagates forward, the feature distribution of the same category gathers gradually in (b) and (c). At the end of the hidden layers, there exist clear margins between the semantic feature distributions of different classes in (d). With the improvement of separability among the feature manifolds, a neuron with lower-precision parameters is able to extract robust features.}
	%\vspace{-0.1in}
	\label{distribution}
\end{figure*}

\subsection{Hybrid Quantization}
\label{sec: Quantization}
%这里表述一下大网络带来的检测增益
%关于量化可以画一张图表示一下这个过程
Powerful deep networks normally benefit from large model capacities but induce high computational and storage costs. Modal quantization is a promising approach to compress deep neural networks, making it possible to be deployed on edge devices. 
%The quantization function is introduced to train a neural network with low-precision weights and activations to compress our network. 
The quantization operator divides the weight into different fixed values by a quantization function which can be regarded as a cluster of convolution kernels in substance. The different scale weights are clustered to a certain value. 

To illustrate this intuition explicitly, a SuperYOLO \cite{zhang2022superyolo} network model which consists of 60 convolutional layers is trained based on the VEDAI dataset. After training, test samples are fed into the model. The convolution weight is firstly clustered into different categories by k-means and then transformed into 2 dimensions by t-SNE \cite{van2008visualizing} to realize the visualization. As shown in Fig. \ref{distribution}, the convolution kernel weight in the (a) $0^{th}$, (b) $26^{th}$ and (c) $52^{nd}$ convolutional layer are clustered in different numbers. In Fig. \ref{distribution} (a), the distance between different categories is relatively far which indicates that the weight distribution is dispersed and complicated in the initial layer. This is due to the fact that the color and texture features, which are detailed and multifarious, are captured in the shallow layer. As the layer propagates forward (Fig. \ref{distribution} (b) and Fig. \ref{distribution} (c)), the convolution weight becomes converging gradually. In other words, the semantic features in the deep layer are more robust and condensed so that with the deepening of the network layers, the clustering categories of weights can be relatively reduced.

\begin{figure}[htb]
	\centering
	\includegraphics[scale=1]{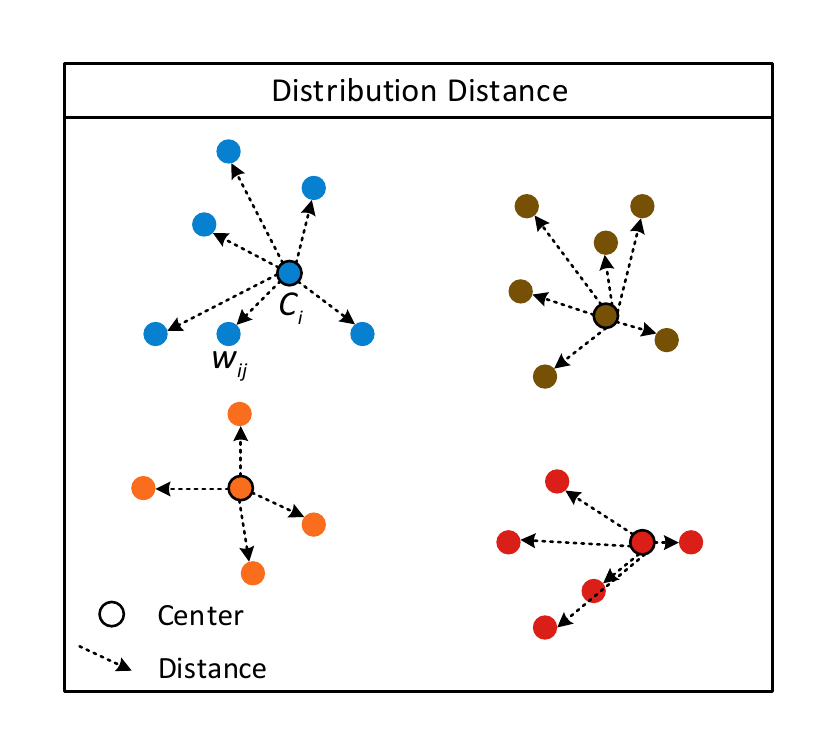}
	\centering
	\caption{The distribution distance of kmeans++ cluster method.}
	\vspace{-0.1in}
	\label{distance}
\end{figure}

Based on this finding, the hybrid quantization idea is introduced to search for the optimal bit width definition in the weight value space. We initially design a hyperparameter $T$ as a threshold to constrain the research space to control the compression rate of the quantization model. The search strategy can be described as:
\begin{equation}
	\begin{split}
		B = &argmax(d(n)) \\
		s.t. &\quad d(n)<T.
	\end{split}
\end{equation}
where the function $d(x)$ denotes the measurement of clustering extent at the $n$ bit width for each convolution layer. This definition aims to find the limited minimum clustering categories (maximum clustering extent) for each layer, hence the smallest quantization model with a minimum bit width is obtained at the preset ratio constraint.
The hybrid quantization of the whole network definitely can be collected as:
\begin{equation}
	B=[B^1,B^2,...,B^l]
\end{equation}
where the $l$ is the total convolution layer and the $B_l$ is the $l^th$ bit width of each layer weight parameter, and the bit width decreases progressively as the network propagates forward.
We utilize the distribution distance defined as follows to determine the final bit width for quantization of the $l^{th}$ convolution layer weight:
\begin{equation}
	\label{distance_equation}
	d^l(n) = \frac{1}{M} \sum\limits_{j = 0}^M {\sum\limits_{i = 0}^{2^n} {{{({w^l_{ij}} - {c^l_i})}^2}} },
\end{equation}
where the $M$ is the total number of kernel weights which correspond to $M = C_{in}\times C_{out}\times K\times K$. The $C_{in}$, $C_{out}$, and $K$ are the input channels, output channels and kernel size of the convolution layer. The whole weight values of each convolution layer complete the kmeans++ algorithm on the different cluster numbers. While the $2^n$ represents the cluster number. $c^l_i$ and $w^l_{ij}$ are the cluster centers and samples, as shown respectively in Fig. \ref{distance}. We set the initial bit width as 8 and then select the superior and adaptive bit width by
\begin{equation}
	\label{selection}
	B^l = \min (n|{d^l(n)} < T) \quad n=b_{min},b_{min}+1,...,8,
\end{equation}
where the $b_{min}$ is a limit of the minimum bit width in the quantization process. When the $d^l(n)$ is smaller than a manual threshold $T$ set in advance, the bit width of the current convolution layer is updated as $B_l$. The activation following this convolution layer keeps the same bit-width.

\begin{figure}[htpb]
	\centering
	\includegraphics[scale=0.6]{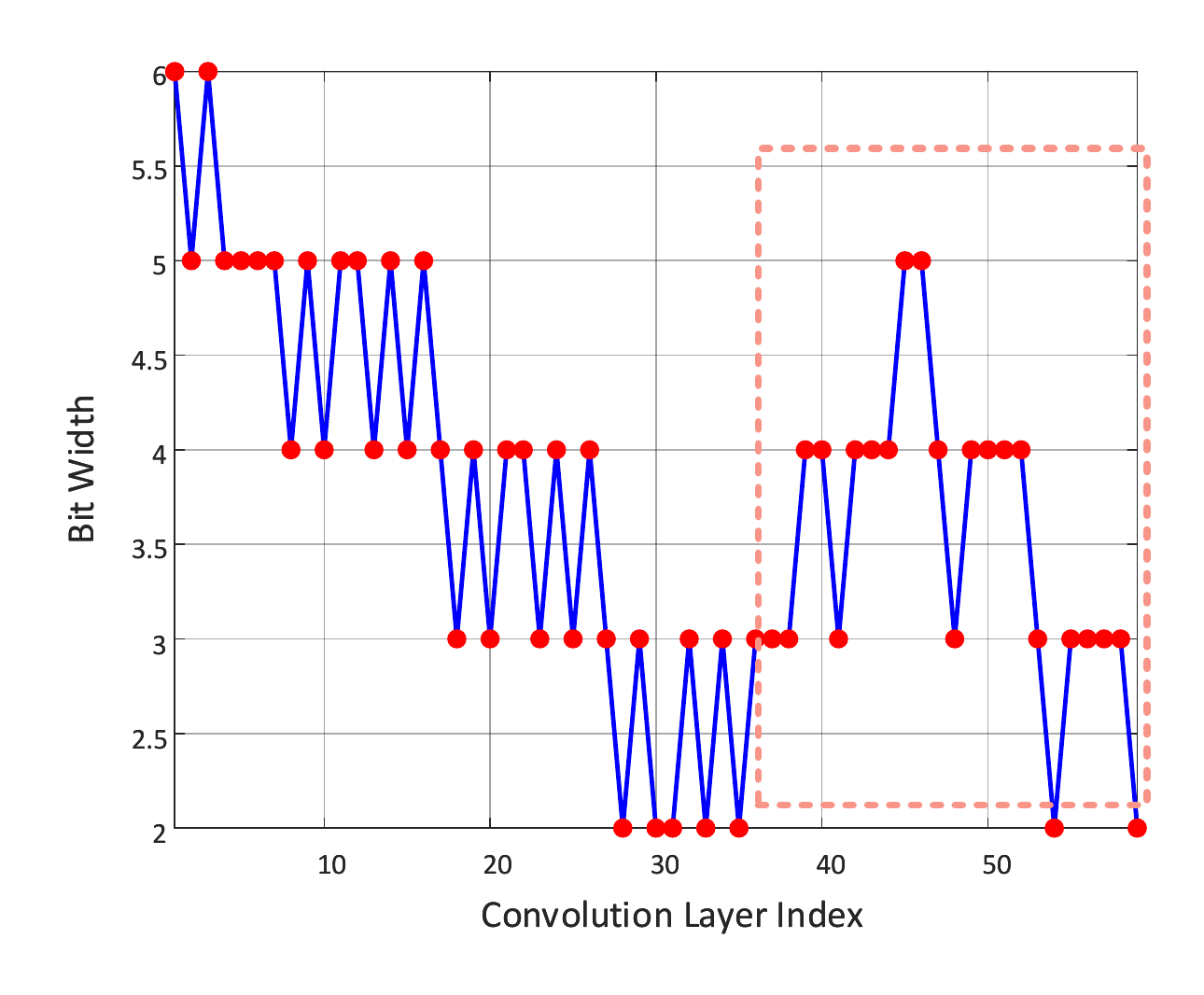}
	\caption{The bit width results of each convolution layer at the threshold $T=50$. The values of bit width progressively decrease with the deepening of the network layer. In addition, the bit width of the convolution layer before the detection module may be relevant large to maintaining more location discrimination information.}
	\vspace{-0.1in}
	\label{bitwidth}
\end{figure}

Take the distance threshold $T=50$ as an example, the Fig. \ref{bitwidth} demonstrates the judgment results of bit width for each convolution layer. It can be indicated that the values of bit width progressively decrease with the deepening of the network layer. In addition, the bit width of the convolution layer before the detection process may be relevantly large to maintain more location discrimination information.
% !TeX spellcheck = en_GB
\begin{algorithm}
	\renewcommand{\algorithmicrequire}{\textbf{Input:}}
	\renewcommand{\algorithmicensure}{\textbf{Output:}}
	\caption{The Hybrid Quantization Method}
	\label{alg:1}
	\begin{algorithmic}[1]
		\Require The weights of $l^{th}$ certain convolution layer ${\bf{W}} \in \mathbb{R}{^{H \times W \times K \times K}}$, The manual distance threshold $T$ and the minimum bit width $b_{min}$ %to adjust the model compression degree by quantization
		\Ensure The bit width of the current convolution layer and activation $B_l$
		\State Initialize the $B_l$ as 8
		\For {$n$ in range ($b_{min}$,8)}
		\State Cluster weights into $2^n$ clusters via the kmeans++ algorithm and then get the centers $c_i$ and samples $w_{ij}$ of the $i^{th}$ cluster.
		\State  Calculate the distribution distance according to Eq. \ref{distance_equation}.
		\State Update the bit width $B_l$ by Eq. \ref{selection}
		\EndFor
		%Judge whether the distribution and aggregation degree meet the requirements.
%		\If {$d<T$}
%            %\If {$N<b_{min}$}
%            %    \State \textbf{return} $b_min$
%            %\Else
%                \State \textbf{return} $N$
%            %\EndIf
%		\ElsIf {$N=8$} 
%    		\State \textbf{return} $8$
%		\EndIf
		\State Complete the quantization for the convolution layer weight and activation by Eq. \ref{convolutionq} and Eq. \ref{activationq}, respectively.

	\end{algorithmic}  
\end{algorithm}

We use a simple-yet-effective quantization method which refers to \cite{zhou2016dorefa} for both weights and activations. The uniform quantization function $q(\centerdot)$ is defined as:
\begin{equation}
	q\left( v,k \right)=\frac{1}{{{2}^{k}}-1}round(({{2}^{k}}-1)v),
\end{equation}
where $v$ is a real number indicating the full-precision (float32) value, $v \in [0,1]$.
the output $q(v,k)$ of quantization function is a $k$ bits real number, $q(v,k) \in [0,1]$.
The quantization calculations of $l_{th}$ convolution layer weight and activation are defined respectively as follows:
\begin{equation}
	\label{convolutionq}
	{{w}^l_{o}}=2q(\frac{\tanh ({{w}^l_{i}})}{2\max (\left| \tanh ({{w}^l_{i}}) \right|)}+\frac{1}{2},B^l)-1,
\end{equation}
\begin{equation}
	\label{activationq}
	{{a}^l_{o}}=q({{a}^l_{i}},B^l).
\end{equation}
The activation $a^l_i$ is the range in $[0,1]$ determined by a bounded activation function while the weight $w^l_i$ is not restricted in a limit boundary. Here, the quantization result of weight $w^l_o$ is the range in $[-1,1]$, and the quantization result of activation $a^l_o$ is the range in $[0,1]$. The Algorithm \ref{alg:1} clarifies the process of the hybrid quantization method. As described in \cite{zhou2016dorefa}, the first and last layers in the network are sensitive to performance during the process of quantization. Based on this intuition, the last detection layer keeps intact to avoid potential degradation of detection performance.

% \begin{equation}
	% s = \frac{{{l_{all}} - l}}{{{l_{all}}}}{\rm{ }}d
	% \end{equation}

\subsection{One-to-one Self-teaching}
\label{sec: Distillation}
Previous mixed quantization approaches pay more attention to the bit-width selection \cite{wang2021generalizable} which costs a lot of resources to obtain the optimal decision. Our hybrid quantization method can make a quick decision with less computation cost. The loss of performance is fixed by the guide of distillation.
In general, previous distillation algorithms are a full precision network, so the network weights are in the same order of magnitude, and the feature maps generated by the teacher network or the student network are similar. However, for the quantitative network, the feature map generated by the quantitative network as a student network will have obvious weight information loss due to the increase of the zero content, resulting in some differences between the feature map of the teacher network and the feature map of the student network, which makes it difficult for the teacher network to directly restrict the quantitative student network from the feature layer. Therefore we proposed an OST to conquer this question.
SCM first calculates the inner connected relationship between the full-precision and quantization network. The distillation switch (DS) chooses the core information between matched student and teacher features by this relationship matrix. We sketch the architecture of self-feature distillation in Fig. \ref{frame}.
%, So we do not distill the information from the middle feature layer, but from the decision layer (referring to the last convolution layer for the detector). We call it the distillation of decision-making layer characteristics.
%通常来说，这种蒸馏算法都是全精度的网络，所以网络权重都是在相同的量级，教师网络或学生网络产生的特征图是相近的。但是对于量化网络来说，量化网络作为学生网络产生的特征图会因为含零量的增加而产生明显的权重信息丢失（能不能画个特征图说明一下），导致教师网络的特征图与学生网络的特征图有一些差异，这就很难让教师网络从特征层去直接约束学生网络（即量化网络），所以我们不从中间的特征层去完成信息的蒸馏，而从决策层（指代的是用于检测器的最后的卷积层）去完成蒸馏任务。我们将其称为决策层特征蒸馏。

% Self-attention is xxx, SA mechanism is a better manner to mine the internal correlation of features, hence alleviating the dependence on external information. Fig. \ref{attention} (a) illustrates the process of the SA module, which is introduced to extra the xxx. In general, the vectors, i.e., Query ($Q$), Key ($K$), and Value ($V$) are obtained by embedding multiplication with three different transformation matrices.

%Self-attention mechanism is a better manner to mine the internal correlation of features, hence alleviating the dependence on external information. In general, the vectors, i.e., Query ($Q$), Key ($K$), and Value ($V$) are obtained by embedding multiplication with three different transformation matrices. The SA layer can be integrally formulated as follows:
%\begin{equation}
%z=Attention(Q,K,V)=softmax(\frac{QK^T)}{\sqrt{d}})V
%\end{equation}
%where the attention weight $a$ can be calculated by $a = softmax(\frac{QK^T)}{\sqrt{d}})$. 
Let $\mathbf{s={s_1,s_2,...,s_l}}$ represent a set of multiscale feature maps for the student network and  $\mathbf{t={t_1,t_2,...,t_l}}$ for the teacher. To calculate the attention map similar to \cite{ji2021show} between the student feature and teacher feature, we define that each teacher feature generates a query $\bf{q_i}$, and each student feature produces a key $\bf{k_j}$:
\begin{equation}
	\mathbf{q_i=W_i \cdot GAP(s_i)},
\end{equation}
\begin{equation}
	\mathbf{k_j=W_j \cdot GAP(t_j)}.
\end{equation}
$\bf{GAP}(\cdot)$ represents a global average pooling. %$a_t$ and $a_s$ are the activation function of the query and key. 
$\bf{W_t}$ and $\bf{W_s}$ are the liner transition parameters for the $i^{th}$ query and the $j^{th}$ key. Then the attention map that reveals the inner relationship between teacher and student features is defined as:
\begin{equation}
	a = ({{\bf{q}} \cdot {\bf{k^T}}})/\sqrt d.
\end{equation}
Here, we introduce the Gumble-Softmax trick \cite{ng2022masked} to convert the values greater than the threshold to 1 and the rest to 0. 
Formally, the decision at $i^{th}$ entry of $a$ is derived in the following way: 
\begin{equation}
	\mathbf{A_{i,k}} = \frac{\exp({\log (\mathbf{a_{i,k}} + \mathbf{G_{i,k}} )/\tau})}{\sum\limits_{j = 1}^K {\exp } ( {\log( \mathbf{a_{i,j}} + \mathbf{G_{i,j}})/\tau } )}
	\quad k = 1,2, \ldots ,K,
\end{equation} 
where K is set as $2$ for binary decision \cite{meng2022adavit} in our case. $\bf G_i$ is the Gumbel distribution. Temperature $\tau$ is used to control the smoothness of $\bf A_i$.
With a better attention map to mining the internal correlation of features, we generate a DS mask that can automatically determine whether to transfer the information from the teacher to the student at the same site in the network.
\begin{equation}
	\alpha  = \bf{Diag}(A).
\end{equation}
where SCM digs out the diagonal elements from the attention map matrix $\bf A$. We devise the self-feature distillation loss as follows:
\begin{equation}
	\mathcal L_{F} = \sum_{i=0}^{m} \alpha_i \Vert \bf{CAP}(t_i) - \bf{CAP}(s_i) \Vert_2,
\end{equation}
where $\bf{CAP}$ represents a channel-wise average pooling. $m$ is the total features utilized for distillation. 
%In addition, the decide-level distillation processed by Kullback-Leibler (KL) divergence:
%\begin{equation}
%\mathcal L_{O} = \bf{KL}(\bf{s_m} \Vert \bf{t_m}).
%\end{equation}
%Here, $\bf{s_m}$ and $\bf{t_m}$ denote the last features which are transferred into the detector to continue detection.
%\begin{equation}
%\mathcal L_{total}= \beta \mathcal L_{F}+ \delta \mathcal L_{O}+ \mathcal L_{dec},
%\label{loss}
%\end{equation}
\begin{equation}
	\mathcal L_{total}= \beta \mathcal L_{F}+ \mathcal L_{dec},
	\label{loss}
\end{equation}
Finally, the distillation loss terms are combined with detection loss and minimized in an end-to-end manner as Eq. \ref{loss}. $\mathcal L_{dec}$ includes the objectness, location, and classification. The hyper-parameter $\beta$ indicates the impact balance between the detection and distillation. 
% In consideration of the spatial and channel domain, we take apart the attention implementation into two parts. The calculation process of relational potential functions about two domains. %are illustrated as Fig. \ref{attentionmap} (a) \cite{wang2018non} and Fig. \ref{attentionmap}. 
% In the spatial domain, the function measures the relationship between any two pixels in the same channel as shown in Fig. \ref{attention} (b). In the channel domain, the function expresses the relationship between any two pixels at the same position as shown in Fig. \ref{attention} (c).
%Then the cosine similarity $S_i$ is calculated between the features xx and xx as:

% !TeX spellcheck = en_US
\section{Experimental Results}
\label{sec: Experiment}
In this section, we evaluate the proposed method on the multimodal dataset for remote sensing object detection and four widely adopted CNNs with different scales. We first demonstrate the experience set up, including the introductions of datasets and networks, implementation details, and evaluation metrics. Then, we report the performance of our method on a dataset in detail, mean average precision and compression ratio are calculated to measure the comprehensive performance in the accuracy and computation cost.

\begin{figure*}[htpb]
	\centering
	\includegraphics[scale=0.9]{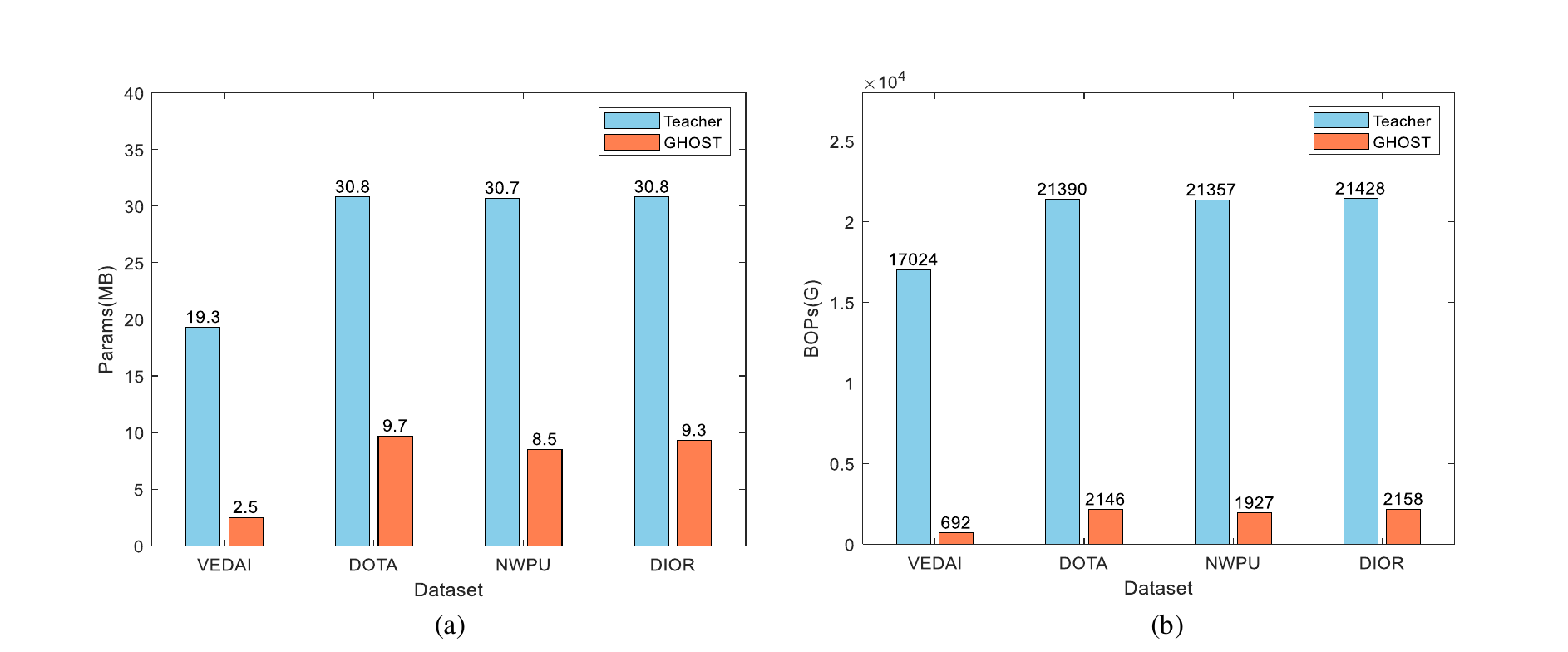}
	\centering
	\caption{Comparisons of the teacher model and lightweight model by parameters and BOPs on the four datasets (VEDAI, DOTA, NWPU, and DIOR). The BOPs and parameters of the lightweight model are smaller, and the inference speed is faster. (a) Params (MB). (b) BOPs}
	%\vspace{-0.1in}
	\label{parabop}
\end{figure*}

\begin{figure*}[htpb]
	\centering
	\includegraphics[scale=0.55]{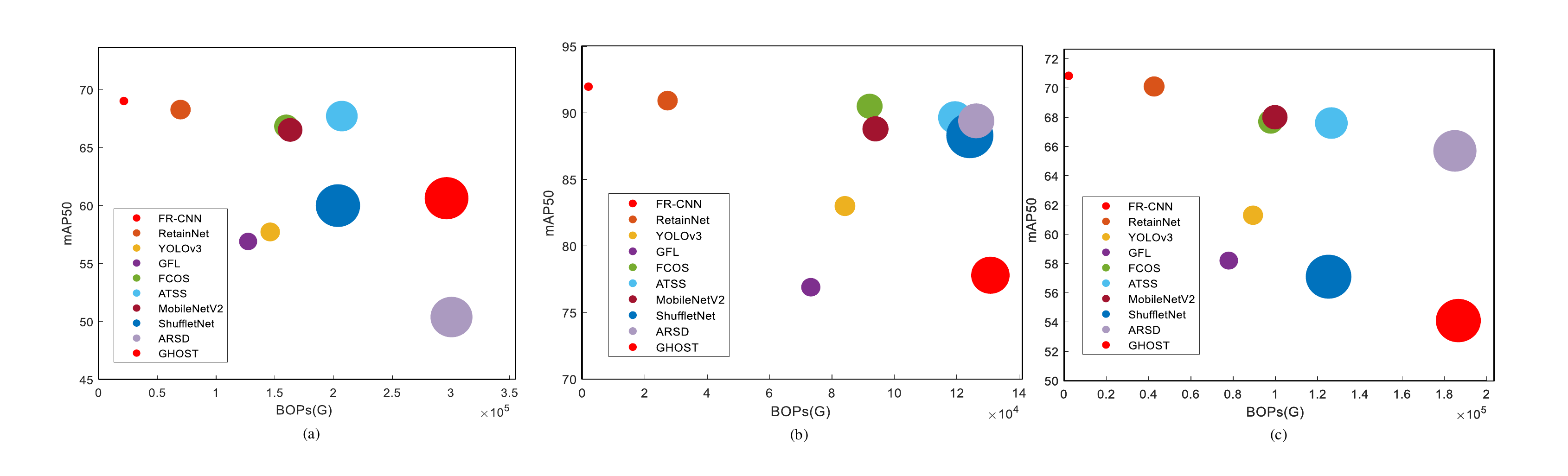}
	\centering
	\caption{Comparison of the efficiency between the current SOTA methods and our method on the three datasets. The bigger size of cycles represents costing more parameters. (a) DOTA. (b) NWPU, and (c) DIOR.}
	%\vspace{-0.1in}
	\label{comparabop}
\end{figure*}

\begin{figure*}[htpb]
	\centering
	\includegraphics[scale=0.5]{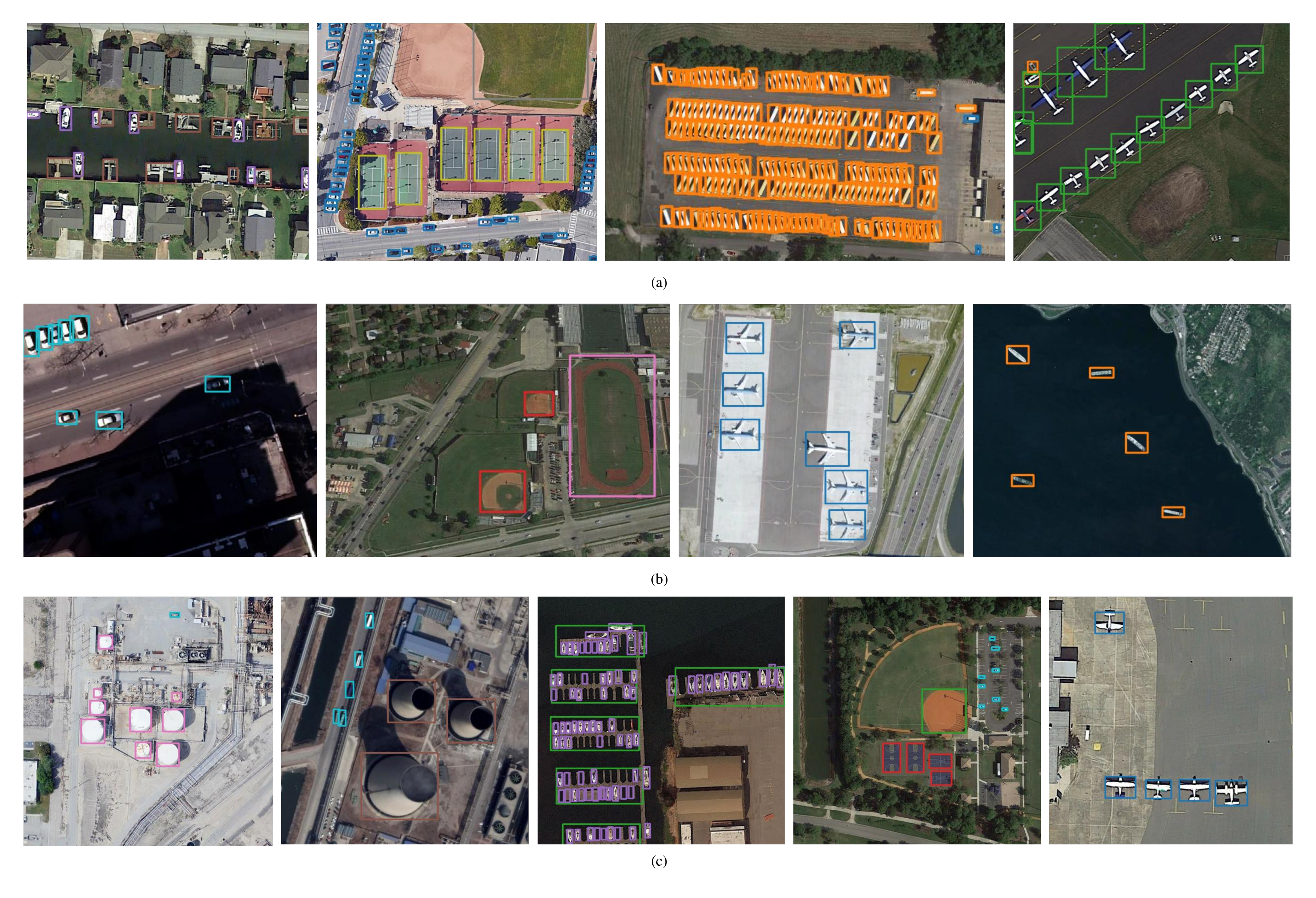}
	\centering
	\caption{Three sets of visualization results. (a) DOTA. (b) NWPU, and (c) DIOR.}
	%\vspace{-0.1in}
	\label{vision}
\end{figure*}

%\subsection{Experiment setup}
\subsection{Dataset Description} 
The publicly available dataset \textbf{VEDAI} \cite{razakarivony2016vehicle}  designed for multimodal remote sensing image object detection is adopted in our experiments. In addition to validation on the multimodal object detection dataset, three single modal datasets (\textbf{DOTA} \cite{xia2018dota}, \textbf{NWPU} \cite{cheng2016learning} and \textbf{DIOR}  \cite{li2020object}) are utilized in experiments to verify the generation of our proposed algorithm.

\textit{1) VEDAI:} The VEDAI dataset consists of 1246 smaller images cropped from the much larger Utah Automated Geographic Reference Center (AGRC) dataset. Each image collected from the same altitude in AGRC has approximately $16,000\times 16,000$ pixels, with a resolution of about $ 12.5cm \times 12.5cm $ per pixel. The main scenes of VEDAI include grass, highway, mountains, and urban areas. The size of the image is fixed to $512 \times 512$.
%All images are the size of $1024\times 1024$ or $512 \times 512 $ pixels. The main eight categories are selected in this paper to complete the object detection task. In this study, $512 \times 512 $ and $1024\times 1024$ images are used to achieve SR branch in object detection assignment. The default image size used in the testing process is $512 \times 512$ unless otherwise specified. The whole experiment is completed on the dataset which is divided into the first fold. %\textcolor{red}{For more information, please refer to \cite{zhang2022superyolo}.}

\textit{2) DOTA:} The DOTA dataset was proposed by Xia et al. in 2018 for object detection of remote sensing. It contains 2806 large images and 188 282 instances, which are divided into 15 categories. The size of each original image is $4000 \times 4000$, and the images are cropped into $1024 \times 1024$ pixels with an overlap of 200 pixels in the experiment. We select half of the original images as the training set, 1/6 as the validation set, and 1/3 as the testing set.  The size of the image is fixed to $512 \times 512$.

\textit{3) NWPU VHR-10:} The dataset of NWPU VHR-10 was proposed by Cheng et al. in 2016. It contains 800 images, of which 650 pictures contain objects, so we use 520 images as the training set and 130 images as the testing set. The dataset contains 10 categories, and the size of the image is fixed to $512 \times 512$. 

\textit{4) DIOR:} The DIOR dataset was proposed by Li et al. in 2020 for the task of object detection, which involves 23 463 images and 192 472 instances. The size of each image is $800 \times 800$. We choose 11 725 images as the training set and 11 738 images as the testing set.

%\begin{table}
%	\small
%	\renewcommand{\arraystretch}{1.3}
%	\centering
%	\setlength{\tabcolsep}{0.8mm}{
%		\caption{Training Strategy}
%		\label{epoch}
%		\begin{tabular}{c|c|c|c|c|c|c|c|c}
%			\toprule[1.2pt]
%			        & \multicolumn{4}{c|}{\textbf{Teacher}}           & \multicolumn{4}{c}{\textbf{GHOST(HQ+OST)}}      \\
%			 \midrule
%			\textbf{Dataset} & \multicolumn{1}{c|}{\begin{tabular}[c]{@{}c@{}}\textbf{Image} \\ \textbf{Size}\end{tabular}} & \multicolumn{1}{c|}{\begin{tabular}[c]{@{}c@{}}\textbf{Batch}\\ \textbf{Size}\end{tabular}}& \textbf{Lr}   & \textbf{Epoch} &\multicolumn{1}{c|}{\begin{tabular}[c]{@{}c@{}}\textbf{Image} \\ \textbf{Size}\end{tabular}} & \multicolumn{1}{c|}{\begin{tabular}[c]{@{}c@{}}\textbf{Batch}\\ \textbf{Size}\end{tabular}} & \textbf{Lr}   & \textbf{Epoch} \\
%			 \midrule[1.2pt]
%			\textbf{VEDAI}   & 512        & 2          & 0.01 & 300   & 512        & 2          & 0.01 & 300   \\
%			\textbf{DOTA}    & 512       & 16          & 0.01 & 100   & 512       & 16         & 0.01 & 100   \\
%			\textbf{NWPU}    & 512        & 8          & 0.01 & 150   & 512        & 8          & 0.01 & 150   \\
%			\textbf{DIOR}    & 512        & 16          & 0.01 & 150   & 512        & 16          & 0.01 & 150   \\
%			\bottomrule[1.2pt]
%	\end{tabular}}
%	%\vspace{-0.1in}
%\end{table}

\begin{table}
	\small
	\renewcommand{\arraystretch}{1.3}
	\centering
	\setlength{\tabcolsep}{2.5mm}{
		\caption{Training Strategy}
		\label{epoch}
		\begin{tabular}{c|c|c|c|c}
			\toprule[1.2pt]
			\textbf{Dataset} &\textbf{Image Size} & \textbf{Batch Size} & \textbf{Lr}   & \textbf{Epoch} \\
			\midrule[1.2pt]
			\textbf{VEDAI}    & 512        & 2          & 0.01 & 300   \\
			\textbf{DOTA}     & 512       & 16         & 0.01 & 100   \\
			\textbf{NWPU}     & 512        & 8          & 0.01 & 150   \\
			\textbf{DIOR}     & 512        & 16          & 0.01 & 150   \\
			\bottomrule[1.2pt]
	\end{tabular}}
	%\vspace{-0.1in}
\end{table}
\begin{table*}
	\small
	\renewcommand{\arraystretch}{1.3}
	\centering
	\setlength{\tabcolsep}{1.1mm}{
		\caption{The comparison result of the tranditonal quantization method and our mixed-bit quantization and we use the same abbreviation in the following sections.}
		\label{quantization}
		\begin{tabular}{cccc|ccccccccccccc}
		\toprule[1.2pt]
        \textbf{Bit Width} & \textbf{T} & \textbf{Max} & \textbf{Min} & Car & Pickup & Camping & Truck & Other & Tractor & Boat & Van & \textbf{mAP50} & \textbf{mAP} & \textbf{Params(MB)} & \textbf{BOPs(G)} \\
        \midrule[1.2pt]
        \textbf{32W32A} & -   & 32 & 32 & 89.20  & 87.10  & 79.50  & 86.80  & 58.20  & 88.00  & 70.30  & 88.30  & \textbf{80.93} & \textbf{50.80} & \textbf{19.30} & \textbf{17023.76} \\
        \midrule
        \textbf{8W8A}   & -   & 8  & 8  & 88.25 & 85.86 & 75.78 & 69.94 & 48.42 & 80.29 & 66.10  & 92.32 & 75.87 & 47.08 & 4.83 & 1201.10  \\
        \rowcolor[HTML]{D9D9D9}
        \textbf{HQ} & 0.1 & 7  & 8  & 86.58 & 83.06 & 69.90  & 76.45 & 69.77 & 76.75 & 69.77 & 99.51 & \textbf{79.57} & \textbf{48.55} & \textbf{4.34} & \textbf{1123.12}  \\
        \midrule
        \textbf{6W6A}   & -   & 6  & 6  & 86.81 & 87.16 & 69.77 & 75.72 & 61.58 & 81.99 & 60.46 & 83.87 & 75.92 & 45.63 & 3.63 & 727.04   \\
        \rowcolor[HTML]{D9D9D9}
        \textbf{HQ}  & 4   & 3  & 8  & 90.94 & 86.57 & 74.45 & 73.35 & 55.92 & 79.70 & 68.63 & 97.84 & \textbf{78.42} & \textbf{46.32} & \textbf{2.49} & \textbf{691.88}   \\
        \midrule
        \textbf{4W4A}   & -   & 4  & 4  & 88.74 & 82.65 & 71.71 & 58.14 & 61.32 & 86.58 & 59.03 & 84.99 & 74.14 & 44.85 & 2.43 & 382.91   \\
        \rowcolor[HTML]{D9D9D9}
        \textbf{HQ}  & 70  & 3  & 8  & 84.72 & 83.41 & 75.13 & 65.39 & 61.41 & 87.77 & 63.52 & 84.73 & \textbf{75.76} & \textbf{46.00} & \textbf{1.87} & \textbf{371.23}   \\
%        \midrule
%        \textbf{2W2A}   & -   & 2  & 2  & 74.19 & 61.04 & 54.97 & 44.88 & 27.61 & 41.03 & 17.50  & 39.40  & 45.08 & 25.21 & 1.22 & 168.69 \\
        
        % 8W8A  & -   & 8 & 8 & 88.25 & 85.86 & 75.78 & 69.94 & 48.42 & 80.29 & 66.1  & 92.32 & 75.87          & 47.08          & 4.83          & 1201.10          \\

        % Mixed & 0.1 & 6 & 8 & 89.00 & 87.28 & 77.30 & 69.80 & 59.79 & 84.12 & 64.91 & 91.77 & \textbf{78.00} & \textbf{46.59} & \textbf{4.12} & \textbf{1107.95} \\
        % \midrule
        % 6W6A  & -   & 6 & 6 & 86.81 & 87.16 & 69.77 & 75.72 & 61.58 & 81.99 & 60.46 & 83.87 & 75.92          & 45.63          & 3.63          & 727.04           \\
        % Mixed & 4   & 3 & 8 & 90.94 & 86.57 & 74.45 & 73.35 & 55.92 & 79.7  & 68.63 & 97.84 & \textbf{78.42} & \textbf{46.32} & \textbf{2.49} & \textbf{691.88}  \\
        % \midrule
        % 4W4A  & -   & 4 & 4 & 88.74 & 82.65 & 71.71 & 58.14 & 61.32 & 86.58 & 59.03 & 84.99 & 74.14          & 44.85          & 2.43          & 382.91           \\
        % Mixed & 70  & 3 & 8 & 84.72 & 83.41 & 75.13 & 65.39 & 61.41 & 87.77 & 63.52 & 84.73 & \textbf{75.76} & \textbf{46.00} & \textbf{1.87} & \textbf{371.23}           \\
        % \midrule
        % 2W2A  & -   & 2 & 2 & 74.19 & 61.04 & 54.97 & 44.88 & 27.61 & 41.03 & 17.5  & 39.4  & 45.08          & 25.21          & 1.22          & 168.69    \\      
        \bottomrule[1.2pt]
	\end{tabular}}
	\vspace{-0.1in}
\end{table*}

\begin{table*}[htpb]
	\small
	\renewcommand{\arraystretch}{1.3}
	\centering
		\caption{The validation result of the self-destillation method in the different model size.}
		\label{distillation}
		\begin{tabular}{ccccc|ccccccccccccc}
		\toprule[1.2pt] 
        \textbf{HQ} & \textbf{T} & \textbf{Max} & \textbf{Min} & \textbf{OST} & Car & Pickup & Camping & Truck & Other & Tractor & Boat & Van & \textbf{mAP50} & \textbf{mAP}  \\
        \midrule[1.2pt]
         $\checkmark$ &0.1 &2&8& & 89.00 & 87.28 & 77.30 & 69.80 & 59.79 & 84.12 & 64.91 & 91.77 & 78.00 & 46.59 &   \\
        \rowcolor[HTML]{D9D9D9}
         $\checkmark$ &0.1 &2&8  &$\checkmark$ &  91.14 & 87.72 & 74.85 & 82.22 & 64.57 & 84.99 & 60.21 & 82.98 & \textbf{78.59} & \textbf{47.14} &  \\
        \midrule
         $\checkmark$ &4 &2&8  && 90.94 & 86.57 & 74.45 & 73.35 & 55.92 & 79.7  & 68.63 & 97.84 & 78.42 & 46.32 &  \\ 
        \rowcolor[HTML]{D9D9D9}
         $\checkmark$ &4 &2&8&$\checkmark$  & 88.58 & 86.16 & 71.84 & 75.18 & 68.57 & 88.14 & 70.44 & 86.80 & \textbf{79.46} & \textbf{48.57}   \\ 
        \midrule
         $\checkmark$ &52 &2&8& & 89.46 & 80.71 & 68.41 & 72.16 & 66.65 & 88.02 & 53.56 & 78.18 & 74.64 & 44.29 &  \\
        \rowcolor[HTML]{D9D9D9} 
         $\checkmark$ &52 &2&8& $\checkmark$ & 88.47 & 83.47 & 71.4  & 72.45 & 60.2  & 83.66 & 66.14 & 89.82 & \textbf{76.99} & \textbf{46.91} \\

        \bottomrule[1.2pt]
	\end{tabular}%}
	\vspace{-0.1in}
\end{table*}

\begin{table*}[htpb]
	\small
	\renewcommand{\arraystretch}{1.3}
	\centering
	\setlength{\tabcolsep}{2.5mm}{
	\caption{The comparison with sota distillation method for detectors on the VEDAI .}
	\label{difdis}
	\begin{tabular}{cc|cccccccccccc}
		\toprule[1.2pt] 
		\textbf{Distillation} & \textbf{HQ} & Car & Pickup & Camping & Truck & Other & Tractor & Boat & Van & \textbf{mAP50} & \textbf{mAP}  \\
		\midrule[1.2pt]
		-    & $\checkmark$ & 89.00 & 87.28 & 77.30 & 69.80 & 59.79 & 84.12 & 64.91 & 91.77 & \textbf{78.00} & \textbf{46.59}  \\ 
		\textbf{ZAQ} \cite{liu2021zero} & $\checkmark$ & 88.04 & 85.86 & 70.51 & 79.45 & 45.16 & 88.14 & 67.76 & 84.03 & 76.12 & 46.48 \\
		\textbf{AFD} \cite{ji2021show} & $\checkmark$  & 88.52 & 85.56 & 71.35 & 73.35 & 58.71 & 89.14 & 59.76 & 80.49 & 75.86 & 45.44 &  \\ 
		\textbf{ReviewKD} \cite{chen2021distilling} & $\checkmark$ & 85.11 & 84.52 & 72.89 & 73.69 & 58.46 & 84.36 & 68.88 & 94.06 & 77.75 & 47.35 \\
		\rowcolor[HTML]{D9D9D9}
		\textbf{OST}  & $\checkmark$ & 91.14 & 87.72 & 74.85 & 82.22 & 64.57 & 84.99 & 60.21 & 82.98 & \textbf{78.59} & \textbf{47.14}  \\
		\bottomrule[1.2pt]
	\end{tabular}}
	\vspace{-0.1in}
\end{table*}

\begin{table}
	\small
	\renewcommand{\arraystretch}{1.3}
	\centering
	\setlength{\tabcolsep}{1.5mm}{
		\caption{mAP comparisons of different $\beta$ value.}
		\label{beta}
		\begin{tabular}{cccccccc}
			\toprule[1.2pt] 
			& \textbf{$\beta$} & 0     & 100   & 200   & 300   & 400   & 500   \\	
			\midrule
			\textbf{\multirow{3}{*}{T}} & 0.1    & 46.59 & 47.17 & 48.26 & 46.72 & \textbf{49.17} & 46.29 \\
			& 4      & 46.32 & 48.57 & 47.05 & 47.96 & 47.29 & \textbf{49.05} \\
			& 52     & 44.29 & 46.91 & 44.20 & 45.14 & 46.48 & \textbf{47.03} \\
			\bottomrule[1.2pt]
	\end{tabular}}
	\vspace{-0.1in}
\end{table}	                    

\begin{table*}[htpb]
	\small
	\renewcommand{\arraystretch}{1.3}
	\centering
	\setlength{\tabcolsep}{1.9mm}{
		\caption{Performance of different algorithms on VEDAI testing set.} %Here ”xWxA ” denotes the bit width for the weight (w) and activation (a)}
		\label{results}
		\begin{tabular}{cccccccccccccc}
			\toprule[1.2pt]
			\toprule[1.2pt]
			\textbf{Method}  & Car   & Pickup & Camping & Truck & Other & Tractor & Boat  & Van   & \textbf{mAP50} & \textbf{mAP}  & \textbf{Params(MB)} & \textbf{BOPs(G)}  \\
			\midrule[1.2pt]
			\textbf{YOLOv3}   & 83.5 & 71.7 & 64.2 & 67.5 & 45.5 & 62.8 & 42.0 & 63.4 & 62.6 & 37.2 & 246 & 50,749 \\
			\textbf{YOLOv4}   & 86.2 & 70.9 & 71.9 & 75.3 & 54.9 & 69.3 & 30.7 & 66.6 & 65.7 & 40.9 & 210 & 39,076 \\
			\textbf{YOLOv5s}  & 81.1 & 71.3 & 70.8 & 66.4 & 58.1 & 67.3 & 27.0 & 55.7 & 62.2 & 34.6 & 28  & 5,427 \\
			\textbf{YOLOv5m}  & 81.6 & 73.9 & 59.0 & 70.0 & 57.2 & 77.7 & 30.5 & 65.5 & 64.4 & 38.2 & 84  & 16,599 \\
			\textbf{YOLOv5l}  & 84.3 & 76.8 & 74.0 & 75.0 & 51.5 & 61.3 & 30.3 & 57.4 & 63.9 & 37.9 & 186 & 37,530 \\
			\textbf{YOLOv5x}  & 84.7 & 66.4 & 66.8 & 72.4 & 65.8 & 67.2 & 29.2 & 58.8 & 63.9 & 37.8 & 349 & 71,301 \\
			\textbf{Teacher}  & 89.2  & 87.1  & 79.5  & 86.8  & 58.2  & 88.0  & 70.3  & 88.3  & 80.93 & 50.80 & 19.3  & 17,024  \\
%			SuperYOLOl  & 91.05 & 88.97 & 73.49 & 83.82 & 69.98 & 82.06 & 68.23 & 95.85 & 81.68 & 50.58 & 125.22 & 125,031.45 \\
%			\midrule[1.2pt]
%			\rowcolor[HTML]{D9D9D9}
%			\textbf{GHOST}  &  88.09 & 83.05 & 72.97 & 73.74 & 68.73 & 86.18 & 74.97 & 89.64 & \textbf{79.67} & \textbf{49.17} & \textbf{4.12} & \textbf{1107.95} \\
			
			\rowcolor[HTML]{D9D9D9}
			\textbf{GHOST}    & 88.82 & 86.06  & 74.33   & 86.96 & 61.28 & 86.00   & 71.76 & 87.26 & \textbf{80.31} & \textbf{49.05} & \textbf{2.5}      & \textbf{692}  \\
%			\rowcolor[HTML]{D9D9D9}
%			\textbf{GHOST}    & 91.06 & 85.46  & 72.93   & 71.63 & 74.21 & 85.31   & 62.46 & 79.56 & \textbf{77.80} & \textbf{47.03} & \textbf{1.46}       & \textbf{377.54} \\
			\bottomrule[1.2pt]
	\end{tabular}}
	\vspace{-0.1in}
\end{table*}

\begin{table*}[htpb]
	\small
	\renewcommand{\arraystretch}{1.3}
	\centering
	\setlength{\tabcolsep}{0.8mm}{
		\caption{Performance of different algorithms on DOTA, NWPU and DIOR testing set.}
		\label{otherdataset}
		\begin{tabular}{cccc|ccc|ccc}
			\toprule[1.2pt]
			\toprule[1.2pt]
            & \multicolumn{3}{c|}{\textbf{DOTA-v1.0}}         & \multicolumn{3}{c|}{\textbf{NWPU}}                    & \multicolumn{3}{c}{\textbf{DIOR}}         \\
            \midrule
			\textbf{Method}         & \textbf{mAP50}          & \textbf{Params(MB)}  & \textbf{BOPs(G) }        & \textbf{mAP50 }               & \textbf{Params(MB)}  & \textbf{BOPs(G)}         & \textbf{mAP50 }    & \textbf{Params(MB)}  & \textbf{BOPs(G)}         \\
			\midrule[1.2pt]
			\textbf{Faster R-CNN}         & 60.64          & 240         & 296,192         & 77.80                & 164         & 130,764         & 54.10     & 240         & 186,572         \\
			\textbf{RetainNet}      & 50.39          & 221         & 300,400         & 89.40                & 145         & 126,228         & 65.70     & 221         & 184,954         \\
			\textbf{YOLOv3}         & 60.00          & 246         & 203,694         & 88.30                & 246         & 124,180         & 57.10     & 247         & 125,153         \\
			\textbf{GFL}            & 66.53          & 76          & 163,000         & 88.80                & 76          & 93,931          & 68.00     & 76          & 99,768          \\
			\textbf{FCOS}           & 67.72          & 126         & 207,001         & 89.65                & 127         & 119,429         & 67.60     & 127         & 126,474         \\
			\textbf{ATSS}           & 66.84          & 75          & 159,754         & 90.50                & 75          & 92,057          & 67.70     & 75          & 97,792          \\
			\midrule
			\textbf{MobileNetV2}    & 56.91          & 41          & 127,221         & 76.90                & 41          & 73,205          & 58.20     & 41          & 77,926          \\
			\textbf{ShuffleNet}     & 57.73          & 48          & 146,022         & 83.00                & 48          & 84,142          & 61.30     & 48          & 89,405          \\
			\midrule
			\textbf{O2-DNet}     & 71.10          & 836          & -         & -                & -          & -          & 68.3     & 836          & -        \\
			\textbf{FMSSD}     & 72.43         &     544     & -         & -                & -          & -          & 69.5     & 544          & -        \\
			\midrule[1.2pt]
			\textbf{Teacher}  & 71.65          & 203          & 291,491               & 93.21                & 127           & 122,234            & 71.7      & 203           & 178,176              \\
			\textbf{ARSD}           & 68.28  \textcolor{green}{-3.37}        & 52          & 69,662          & 90.92 \textcolor{green}{-2.29}          & 46          & 27,289          & 70.10    \textcolor{green}{-1.6}    & 52          & 42,598          \\
			\midrule
			\rowcolor[HTML]{D9D9D9}
			\textbf{Teacher} & 69.99 & 30.8 & 21,390 & 93.30   & 30.7 & 21,357 &  71.95 & 30.8 & 21,428 \\
			\rowcolor[HTML]{D9D9D9}
			\textbf{GHOST}          & \textbf{69.02 \textcolor{green}{-0.97}}     &  \textbf {9.7}         &    \textbf{2,146}             & \textbf{91.97 \textcolor{green}{-1.33}} & \textbf{8.5}  & \textbf{1,927}  & \textbf{71.53  \textcolor{green}{-0.4}} &      \textbf{9.3}       &          \textbf{2158}       \\
%			\textbf{GHOST}          &                &             &                 & \textbf{}            & \textbf{}   & \textbf{}       & \textbf{} & \textbf{}   & \textbf{}       \\
%			\textbf{GHOST}          &                &             &                 & \textbf{}            & \textbf{}   & \textbf{}       & \textbf{} & \textbf{}   & \textbf{}        \\
			\bottomrule[1.2pt]
	\end{tabular}}
	%\vspace{-0.1in}
\end{table*}

\subsection{Implementation Details}
\subsubsection{Networks}
To demonstrate superior performance, SuperYOLO \cite{zhang2022superyolo} is tested as a teacher model with our method. For the multimodal VEDAI dataset, the number of convolution layers is 47 including one detection layer on the small scale.   For a single modal dataset (DOTA, NWPU, and DIOR), the number of convolution layers is 61 including three detection layers on the small, medium, and large scale. 
To verify the superiority of the GHOST proposed in this paper, we selected 12 generic methods for comparison: 

one-stage algorithms (\textbf{YOLOv3} \cite{redmon2018yolov3}, \textbf{YOLOv4} \cite{bochkovskiy2020yolov4}, \textbf{YOLOv5} \cite{yolov5}, \textbf{SuperYOLO},  \textbf{FCOS} \cite{tian2019fcos},  \textbf{ATSS} \cite{zhang2020bridging}, \textbf{RetainNet} \cite{lin2017focal}, \textbf{GFL} \cite{li2020generalized});

two-stage method (\textbf{Faster R-CNN} \cite{ren2015faster});

lightweight models (\textbf{MobileNetV2} \cite{sandler2018mobilenetv2} and \textbf{ShuffleNet} \cite{zhang2018shufflenet}); 

distillation-based methods (\textbf{ARSD} \cite{yang2022adaptive}); 

remote sensing designed approaches (\textbf{FMSSD} \cite{2020FMSSD} and \textbf{O2DNet} \cite{WEI2020268}).

\subsubsection{Training Strategy}
Our proposed framework is implemented in PyTorch and runs on a workstation with an NVIDIA A100-SXM4-80GB GPU. We also use different training strategies for different datasets,  and the detail is illustrated in TABLE \ref{epoch}. In addition, data is augmented with Hue Saturation Value (HSV), multi-scale, translation, left-right flip, and mosaic. The augmentation strategy is canceled in the test stage. The standard Stochastic Gradient Descent (SGD) is used to train the network with a momentum of 0.937, weight decay of 0.0005 for the Nesterov accelerated gradients utilized, and a batch size of 2. The learning rate is set to 0.01 initially.  All the baseline training process is completed from scratch without any pre-trained model while the GHOST is carried on the baseline model.
%The ablation experiment is completed on 300 epochs for a fast invalidation. The entire training process involves 400 epochs to obtain high accurate fully-precision network and provide sufficient time for quantization to recover performance.
In the test stage, the IoU threshold of non-maximum suppression is 0.6 on NWPU VHR-10 and VEDAI, and it is 0.4 on DOTA and DIOR.

\subsubsection{Evaluation Metric}
For the detection result, the IoU is defined as the ratio of the intersection and union of two boxes. During the evaluation, according to the IoU of predicted boxes and ground truths, each sample will be assigned attributes: true positive (TP) for correctly matching, false positive (FP) for wrongly predicting the background as an object, and false negative (FN) for the undetected object. 
During the evaluation, all the detection boxes are sorted in order of confidence score from high two low and then traversed. In the traversed process, the calculations of the precision and recall metrics can be defined as:
\begin{equation}
	Precision=\frac{TP}{TP+FP},
\end{equation}
\begin{equation}
	Recall=\frac{TP}{TP+FN}.
\end{equation}
The precision and recall are correlated with the commission and omission errors, respectively. The AP values use an integral method to calculate the area enclosed by the Precision-Recall curve and coordinate axis of all categories. Hence, the AP can be calculated by
\begin{equation}
	AP=\int_{0}^{1}{p(r)dr},
\end{equation}
where $p$ denotes Precision, $r$ denotes Recall. The mAP is a comprehensive indicator obtained by averaging APs for all classes. 
%Moreover, Parameters are the number of network parameters, representing the size of the model. Floating Point Operation (FLOP) measures the amount of computation and the complexity of the model.
Moreover, we choose Bit-Operations (BOPs) count \cite{wang2020differentiable} and parameters to measure the compression performance. The Bops of convolution are calculated as:
\begin{equation}
	BOPs_l = c_{l-1}\times c_{l} \times w_l \times h_l \times k_w \times k_h \times b_{w,l} \times b_{a,l-1}.
\end{equation}
The $h_l$, $w_l$, and $c_l$ are the with, height, and several channels of the $l-th$ layer output feature map, respectively. $b_{w,l}$ and $b_{a,l}$ denote $l-th$ layer weight and activation bit-weight.
The parameters (params) are defined as: 
\begin{equation}
	params = \frac{c_{l-1} \times c_{l} \times k_h \times k_w \times b_{w,l}}{8bit} (B).
\end{equation}
%\subsection{Measure the Efficiency of the Model}
%Frames Per Second (FPS) is the number of frames detected per second, which represents the running speed of the deep learning model. We input $1024 \times 1024$ pixel resolution images of the entire test set and calculate the average number of images processed per second as FPS. Parameters are the number of network parameters, representing the size of the model. Floating Point Operation (FLOP) measures the amount of computation and the complexity of the model.
%\subsection{Ablation Study}

\subsection{Ablation Study}
%\label{sec:Ablation Study}
In this section, we conduct the ablation experiments of our GHOST framework. We explore how each module (HQ and OST) compresses the model and promotes the performance of the small student model. Besides, the experiments of different distillation algorithms and distillation hyperparameters' optimization are carried out. We conduct ablation experiments on the dataset of VEDAI for object detection. 

\textbf{Validation of HQ:}
Distribution distance hybrid quantization can integrate device n-bit settings for the network, so we experiment with such variations of integrated hybrid n-bit quantization. To analyze the performance of differences, we compare fixed  DoReFa-Net \cite{zhou2016dorefa} and various hybrid DoReFa-Net quantization methods on the SuperYOLO detection network. As illustrated in TABLE \ref{quantization}, the HQ is the proposed hybrid quantization method, and the $\cdot W \cdot A$ presents the traditional unified bit width quantization algorithm except $32W32A$ represents the full precision network. Max and Min denote the maximum and minimum bit width in the quantization model. Hybrid quantization enables the quantization model to preserve the significant information to achieve the minimal accuracy loss possible.
As shown in TABLE \ref{quantization}, the hybrid quantization module accomplishes the optimal consequence and detection accuracy has reached $79.57\%$, $78.42\%$, and $75.76\%$ respectively, which are more $3.7\%$, $2.5\%$, and $1.62\%$ than fixed quantization at the different computation orders. The hybrid quantization achieves better accuracy in the VEDAI dataset than the accuracy of fixed quantization costing fewer computation resources (parameters and BOPs). 

\textbf{Effect of OST Module:}
After the HQ module has been added to the network, we also adopt one-to-one self-teaching within the three-quantization scale. Table \ref{distillation} is based on SuperYOLO which is used as the teacher network and student network simultaneously. The experiments are carried out on the VEDAI dataset.  The OST module enables the typical detection network to recover the performance of the quantization detection network improved by $0.59\%$, $1.04\%$, and $2.35\%$, respectively no matter what kind of bit width.

\textbf{Comparison with the SOTA Distillation Method:}
In addition, Table \ref{difdis} shows the comparisons between the proposed OST module and existing distillation frameworks. OST achieves superior performance in the field of remote sensing under the premise of the same computation cost while the ZAQ, AFD, and ReviewKD lead to an accuracy degradation of the quantization network. And also it can be proved that OST distillation is a benefit for the guidance between the full-precision model and the quantization model.

\textbf{Hyperparameters Optimization:}
As presented in Section \ref{sec: Distillation}, $\beta$ is the distillation weights of the OST module, so we compare the performance of the distillation in the different weights which are shown in TABLE \ref{beta}. We conduct the hyperparameter experiment on the VEDAI dataset on GHOST to find the best $\beta$. As shown in the TABLE \ref{beta} shows, the model reaches the best when $\beta=400$ at the $T=0.1$ and when $\beta=400$ at the $T=4$ or $T=52$.

\textbf{Lightweight Analysis:}
Owning to the GQSD design idea, our student model is very lightweight. We compare the GHOST and the teacher model in terms of parameters and BOPs on the four datasets. As Fig. \ref{parabop} shows, GHOST has smaller model parameters, and fewer BOPs compared with the teacher model no matter in which dataset. Hence, our compression strategy for the lightweight model is more practical to be deployed on intelligent terminals.

\subsection{Comparison with  the SOTA Detectors}
In this part, we compare our lightweight model with other classic heavy object detection methods. As shown in TABLE \ref{results} and TABLE \ref{otherdataset}. Experiments on the four datasets prove the efficiency and efficacy of the GHOST framework. Not only does our lightweight have higher accuracy but also it has strong information retention capability under extreme model compression.

\textit{1) VEDAI:} Our GHOST achieves 80.31\% mAP50 compared with other detectors, surpassing the one-stage series mentioned in TABLE \ref{results}. Our model achieves the lowest model parameters (2.5 MB) and BOPs (692 G).

\textit{2) DOTA:}  As presented in TABLE \ref{otherdataset}, our GHOST achieves the optimal detection result (69.02\% mAP50) and the model parameters (9.7 MB) and BOPs (2,146 G) are much smaller than other SOTA detectors regardless of the two-stage, one-stage, anchor-free or distillation-based method. We also compare two detectors designed for remote sensing imagery such as FMSSD \cite{2020FMSSD} and O2DNet \cite{WEI2020268}. Although these models have a close performance with our lightweight model, the huger parameters and BOPs seem to be a massive cost in computation resources. Hence, our model has a better balance in consideration of detection efficiency and efficacy. Compare to the distillation-based method ARSD (-3.37\%), the GHOST obtains an even smaller accuracy gap (-0.97\%) between the student network and the teacher network. It demonstrates that our GHOST method can transfer sufficient knowledge to guide the learning of the student model and the misunderstanding between both models can be reduced by the SCM training strategy.

\textit{3) NWPU:}  We compare the results of our method with other approaches on the NWPU dataset. As shown in TABLE \ref{otherdataset}, our GHOST obtains the best result (91.97\% mAP50) with the smallest amount of model parameters (8.5 MB) and the fewer BOPs (1,927 G). 

\textit{4) DIOR:} As illustrated in TABLE \ref{otherdataset}, our GHOST achieves the optimal detection result (69.02\% mAP50) and the model parameters (9.3 MB) and BOPs (2,158 G) are much smaller than other SOTA detectors regardless of the two-stage, one-stage, anchor-free lightweight, distillation-based methods. It reveals the strong ability to compress models and the power capacity of object detection in remote sensing imagery. The accuracy of the student GHOST is only a bit less 0.4\% than teacher network, compared with ARSD (1.6\%).

In order to show the performance of our algorithm more intuitively, We compare the detection accuracy, parameters, and BOPs of various algorithms in Fig. \ref{comparabop}.  It can be obvious that GHOST has a better trade-off between performance and lightweight.  The visualization results on the three datasets are illustrated in Fig. \ref{vision} in which we can see that the GHOST can have an outstanding detection of objects at different scales.

% !TeX spellcheck = en_US
\section{Conclusion}
\label{sec: Conclusion}

In this paper, we propose a GHOST framework for a lightweight object detection method in remote sensing imagery. We first design a guided quantization self-distillation structure which is not only a training technique to preserve model performance but also a method to compress and accelerate models. Although most of the previous research focuses on knowledge transfer among different models, we believe that inside distillation is also very promising.  Secondly, we propose a hybrid quantization that captures the optimal bit width selection based on an adaptive way in the weight value research space to break the limit of the fixed quantization model accuracy. Thirdly, the proposed one-to-one self-teaching module gives the student network of self-judgment through a switch control machine that accurately handles the knowledge transformation. It can dynamically discriminate the wrong guidance and mine the effective knowledge from the teacher. The experiments based on the VEDAI, DOTA, NWPU, and DIOR datasets certify that our GHOST achieves SOTA performance compared with other detectors. It can well balance the tradeoff between accuracy and specific resource constraints.

%\end{frontmatter}

%\linenumbers

%% \linenumbers

%% main text
%\IEEEpeerreviewmaketitle

\ifCLASSOPTIONcaptionsoff
\newpage
\fi
{
	\bibliographystyle{IEEEtran}
	\bibliography{reference}
}

\end{document}